\documentclass[10pt,journal,twocolumn]{IEEEtran}
\ifCLASSOPTIONcompsoc
  \usepackage[nocompress]{cite}
\else
  \usepackage{cite}
\fi

\ifCLASSINFOpdf
\else
\fi

\usepackage{times}
\usepackage{epsfig}
\usepackage{graphicx}
\usepackage{amsmath}
\usepackage{amssymb}
\usepackage{eucal,bibspacing}
\usepackage{cs}
\usepackage{mathsymb}
\usepackage{textcomp}
\usepackage{verbatim}
\usepackage{url}
\usepackage{multirow}
\usepackage[super]{nth}
\usepackage{colortbl}
\usepackage{color}
\usepackage{pbox}
\usepackage{xcolor}

\usepackage{wrapfig}
\usepackage{caption}
\usepackage[linesnumbered,algoruled,boxed,lined]{algorithm2e}

\def\be{{\boldsymbol e}}
\def\bx{{\boldsymbol x}}

\let\savedalgorithm\algorithm
\let\savedendalgorithm\endalgorithm

\usepackage{algorithm}

\newcommand{\eg}{\textit{e.g., }}

\begin{document}

\title{Deep Co-attention based Comparators For Relative Representation Learning in Person Re-identification}

\author{Lin Wu, Yang Wang, Junbin Gao, Dacheng Tao, \textit{Fellow, IEEE}
\IEEEcompsocitemizethanks{\IEEEcompsocthanksitem L. Wu is with ITEE, The University of Queensland, St Lucia 4072, Australia. Email: lin.wu@uq.edu.au.

Y. Wang is with Dalian University of Technology, 116024, Dalian, China. Email: yang.wang@dlut.edu.cn, Corresponding author

J. Gao is with Discipline of Business Analytics, The University of Sydney Business School, The University of Sydney, NSW 2006, Australia. Email: junbin.gao@sydney.edu.au.

D. Tao is with the UBTech Sydney Artificial Intelligence Institute, The University of Sydney, Darlington, NSW 2008, Australia, and also with the Faculty of Engineering and Information Technologies, School of Information Technologies, The University of Sydney, Darlington, NSW 2008, Australia. Email: dacheng.tao@sydney.edu.au. \protect\\

}
}

\IEEEtitleabstractindextext{%
\begin{abstract}
Person re-identification (re-ID) requires rapid, flexible yet discriminant representations to quickly generalize to unseen observations on-the-fly and recognize the same identity across disjoint camera views. Recent effective methods are developed in a pair-wise similarity learning system to detect a fixed set of features from distinct regions which are mapped to their vector embeddings for the distance measuring. However, the most relevant and crucial parts of each image are detected independently without referring to the dependency conditioned on one and another. Also, these region based methods rely on spatial manipulation to position the local features in comparable similarity measuring. To combat these limitations, in this paper we introduce the Deep Co-attention based Comparators (DCCs) that fuse the co-dependent representations of the paired images so as to focus on the relevant parts of both images and produce their \textit{relative representations}. Given a pair of pedestrian images to be compared, the proposed model mimics the foveation of human eyes to detect distinct regions concurrent on both images, namely co-dependent features, and alternatively attend to relevant regions to fuse them into the similarity learning. Our comparator is capable of producing dynamic representations relative to a particular sample every time, and thus well-suited to the case of re-identifying pedestrians on-the-fly. We perform extensive experiments to provide the insights and demonstrate the effectiveness of the proposed DCCs in person re-ID. Moreover, our approach has achieved the state-of-the-art performance on three benchmark data sets: DukeMTMC-reID \cite{DukeMTMC}, CUHK03 \cite{FPNN}, and Market-1501 \cite{Market1501}.
\end{abstract}

\begin{IEEEkeywords}
Attention Models, Person Re-identification, Co-attention, Relative Representations
\end{IEEEkeywords}}

\maketitle

\IEEEdisplaynontitleabstractindextext

\IEEEpeerreviewmaketitle

\section{Introduction}\label{sec:intro}

\IEEEPARstart{P}{erson} re-identification (re-ID) is a critical task in visual surveillance that aims to associate individuals across disjoint cameras at different times. It is of great security interest and can be used for various surveillance applications, such as facilitating cross-camera tracking of people and understanding their global behavior in a wider context \cite{Time-delayed}. However, matching pedestrians with visual appearance is non-trivial owing to the visual ambiguities, caused by illuminations, viewpoints and human pose changes. Recent studies focus on computing the representations over the human body regions and achieve notably improved recognition results \cite{Multi-channel-part,S-LSTM,SpindleNet,Part-Aligned,Context-body,What-and-where}. The key idea is to detect the discriminative human body regions/parts from which local representations are produced for corresponding similarities. Nonetheless, region-based methods still have some technique limitations. First, precise localization over each informative region is very difficult and existing region based methods resort to manually designed horizontal windows \cite{GatedCNN,S-LSTM,LOMOMetric,SimilaritySpatial}. On the other hand, the local feature learning is performed individually on each single image and extracted features are from a fixed set of detected regions.

Recently, attention based deep neural networks \cite{Diversity-attention,Two-attention,LinTcyb18,ShowAttendTell} are studied with attempt to be applicable to a range of tasks including scene generation \cite{DRAW} and fine-grained recognition \cite{Diversity-attention}. In person re-ID, some attention based approaches have been proposed \cite{Com-attention,What-and-where,Deepadaptive} to focus on the most salient regions which help discriminate individuals across camera views. However, these attention based approaches simply perform a single-pass structure sequentially on each pedestrian image without contextualizing feature dependence across images. In fact, matching paired images is to correlate their visual structure through an accumulation of attention on them. Also, it is demonstrated that flexible representations that can capture the interactions between the paired images are very crucial to re-ID \cite{GatedCNN,What-and-where}. Finally, potential regions proposed by attention still need to be spatially positioned to allow comparable matching. These limitations hinder the application of these methods in practical re-ID systems that demand a rapid learning competence to quickly attend to distinct regions across images and judge their similarity.

In this paper, we present a novel neural network architecture to learn \textit{relative representations} for person re-ID, that is able to quickly generalize to unseen observations and determine their similarity by attending to distinct regions \textit{relative} to each paired image in comparison. We are motivated by the observation that each pair should have different regions to be compared for the sake of distinguishing the pair to be the same or not. In this sense, regional features are conditioned on each other, namely concurrent w.r.t the pair to be compared on-the-fly. The other observation is given concurrent regions across images the human perform similarity judgement by alternating these features with accumulation to determine the similarity. This natural accumulation on comparable features should be captured to eliminate the unnecessary spatial manipulation.

The proposed model is named Deep Co-attention based Comparators (DCCs) in the rest of this paper. The networks have ingredients of co-dependent region detection and a selective spatial attention mechanism that resemble the foveation of human eyes to sequentially compare persons conditioned on each other. It naturally correlates paired images to be compared step by step, and learns to have the ability to selectively attend to parts of the person images while ignoring irrelevant parts. Some recent one-shot/few-shot learning methods are demonstrated to be viable to person re-ID \cite{One-shot-RE-ID,t-LRDC,Cross-GANs} in terms of adaption to different domain samples. Also some unsupervised approaches are presented to have promising generalization on recognizing unseen observations \cite{UMDL,L1-graph,OL-MANS,PUL}. However, these methods are still limited in inherent technique bottlenecks. First, they don't embed similarity learning into their feature learning. The separation on feature and similarity learning is unlikely to produce optimal feature representations \cite{PersonNet,JointRe-id}. On the other hand, without the aid of supervision, unsupervised methods are unable to produce discriminative features, and thus not competing to supervised alternatives.

\subsection{How do human compare objects?}

\begin{figure}[t]
\centering
\begin{tabular}{c}
\includegraphics[height=5cm]{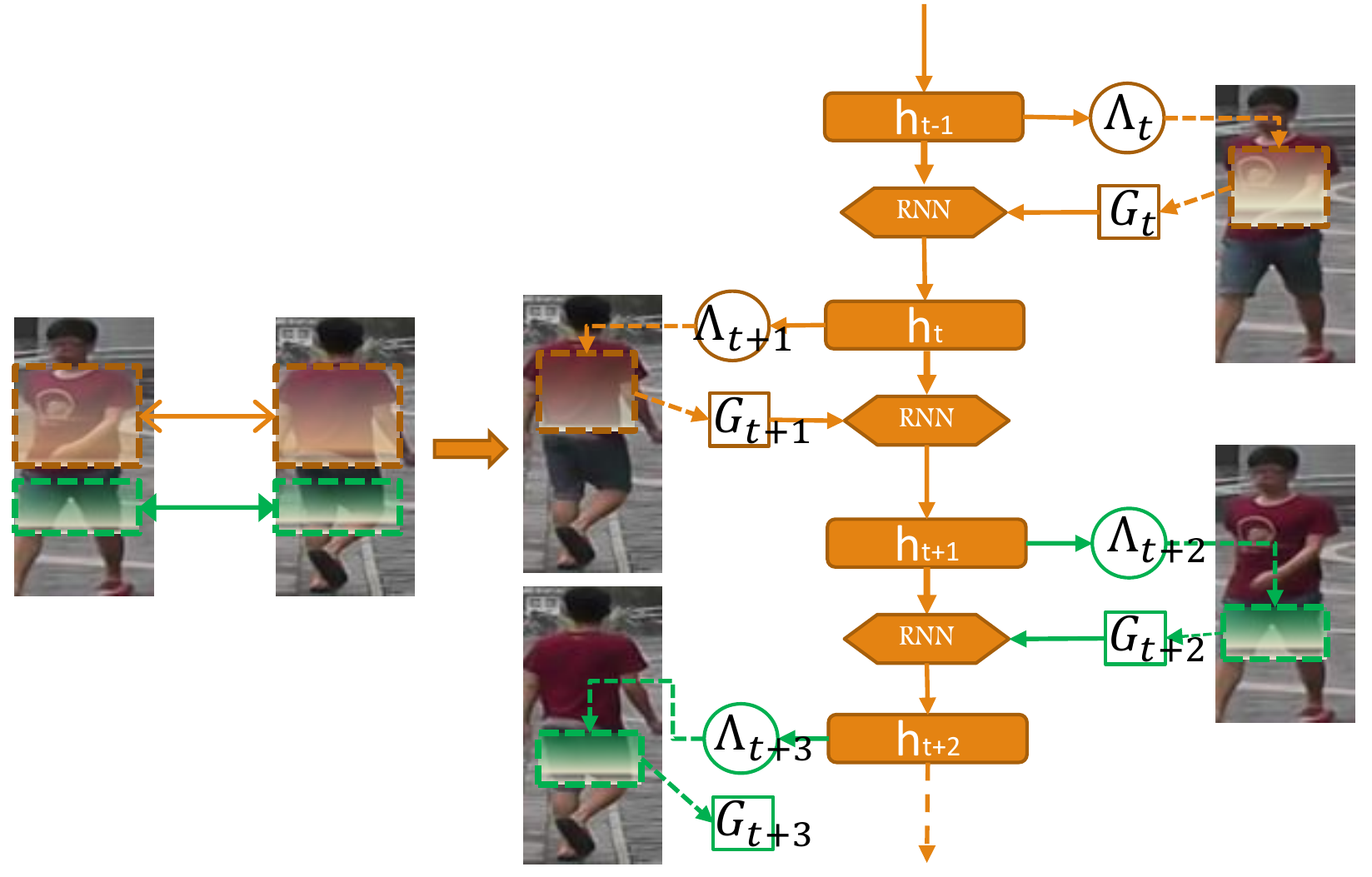}
\end{tabular}
\caption{The sketch of a DCC on comparing two images. The co-dependent regions are detected and encoded by the deep co-attention encoder (See Section \ref{ssec:coattention}). These regions are alternatively and repeatedly attended, and the dynamic discriminant representations are learned by fusing the information across the input paired images (See Section \ref{ssec:iterative-comparator}). In this model, the similarity is compared by making an observation in one image conditioned on the observations made so far in both images.}
\label{fig:comparator}
\end{figure}

The proposed DCCs are inspired by our interpretation of how human generally compare a pair of visual objects without prior knowledge (See Fig.\ref{fig:comparator}). When a human is asked to compare two pedestrians in images and estimate their similarity, the person would first determine which parts of one image should be observed while referring to other image (\textit{co-dependence}), and then he performs the comparison by repeatedly looking back and forth between the two images with fixation on attentive regions (\textit{recurrent comparison}).

In person re-ID, the dominating similarity estimation system based on deep learning is derided from the Siamese pipeline \cite{JointRe-id,PersonNet,FPNN,GatedCNN}. In this major methodology, they essentially perform the similarity estimation based on local distinct regions. Towards this end, existing methods learn to detect local regions robust to visual variations which can be further mapped to their embeddings with corresponding distance measuring. However, detecting useful regions is carried out independently on a single image and the similarity is learned only after the completion of detection. In other words, the Siamese way is most likely to fuse the compared embeddings only in the last stage and at a higher abstract level. Moreover, spatial manipulation is additionally needed to allow the extracted local features compared in the context of semantic meaning.

\subsection{Our approach and contributions}

In this paper, we present a novel deep learning approach to person re-ID on-the-fly to produce relative representations with respect to each unseen paired images and their similarity measure. Our model is distinct from existing state-of-the-arts in the merits of resembling the human foveation in fusing the most relevant patterns into their similarity learning by the iterative recurrence comparator. Hence, data-dependent features can be learned and the spatial manipulation is alleviated. The major contributions of this paper can be summarized as follows.
\begin{itemize}
\item We introduce a novel deep architecture to person re-ID that is able to concurrently detect the most distinct patterns from paired images and fuse them into the similarity learning.
\item The proposed Deep Co-attention based Comparators (DCCs) proceed with two-stream convolutional feature maps corresponding to paired images, which are correlated by the co-dependent encoder to render respective features contextual aware of each other.
\item Insightful analysis to our approach and extensive experiments on benchmark data sets are provided in this paper.
\end{itemize}

The rest of this paper is organized as follows. Section \ref{sec:related} reviews related works from person re-identification and recent attention based models. We detail the proposed DCCs in Section \ref{sec:approach} wherein the architecture of the network and each component are described. Section \ref{sec:exp} reports extensive experiments over three benchmarks, and the paper is concluded in Section \ref{sec:con}.

\section{Related Work}\label{sec:related}

\subsection{Person Re-identification}

Existing person re-ID pipelines can be roughly categorized into two streams: feature learning and similarity metric learning. In the first pipeline, deep feature learning from local regions/patches directs the research efforts. For instance, some patch-level matching methods with spatial layout are proposed \cite{GenerativeSaliency,Zhao2013SalMatch,eSDC,Farenzena2010Person} which segment images into patches and perform patch-level matching with spatial relations. Methods in \cite{MatchTemplate,Farenzena2010Person} separate images into semantic parts (\eg head, torso and legs), and measure similarities between the corresponding semantic parts. However, these methods assumes the presence of the silhouette of the individual and accuracy of body parser, rendering them not applicable when body segmentations are not reliable. To avoid the dependency on body segments, saliency-based approaches \cite{Zhao2013SalMatch,eSDC} are developed to estimate the saliency distribution relationship and control path-wise matching process.
Some metric learning approaches \cite{SimilaritySpatial,LocalMetric,LADF,YangMM2015,LOMOMetric,Bar-shape,Yang-TIP15,Yang-TNNLS17,Yang-TNNLS18,Yang-TIP17} make attempts to extract low-level features from local regions and perform local matching within each subregions. They aim to learn local similarities and global similarity, which can be leveraged into an unified framework. Despite their effectiveness in local similarity measurement with some spatial constraints, they have limitations in the scenarios where corresponding local regions are roughly associated.

With the resurgence of Convolutional Neural Networks (CNNs) in image classification \cite{AlexNet}, a number of deep similarity learning methods based Siamese CNN architecture \cite{FPNN,JointRe-id,PersonNet,Multi-channel-part,SI-CI,GatedCNN} are proposed for person re-id with the objective of simultaneously learn discriminative features and corresponding similarity metric. However, current networks extract fixed representations for each image without consideration on concurrent local patterns which are crucial to discriminate positive pairs from negatives. In contrast, we present a model to learn flexible representations from common local patterns which are robust against cross-view transformations. S-CNN \cite{GatedCNN} has some sharing with us in emphasizing finer local patterns across pairs of images, and thus flexible representations can be produced for the same image according to the images they are paired with. However, their matching gate is to compare the local feature similarities of input pairs from the mid-level, which is unable to mediate the pairwise correlations to seek joint representations effectively. Moreover, S-CNN \cite{GatedCNN} manually partitions images into horizontal stripes, and this renders S-CNN unable to address spatial misalignment. The other work close to us is the end-to-end comparative attention network (CAN) \cite{Com-attention} that learns to selectively focus on parts of paired person images, and adaptively compare their appearance. However, CAN needs to generate multiple glimpses from the same image to localize different parts in which spatial relationship is not explicitly modeled. Moreover, CAN is using a standard visual attention model which is more likely to generate similar attention maps at different time steps \cite{Diversity-attention}, and thus smaller regions cannot be discovered to differentiate visually similar persons. Another work in \cite{What-and-where} has introduced multiplicative integration gating mechanism to learn joint representations attentively from common local finer patterns. However, their representation learning is subject to the spatial recurrence to address the spatial misalignment. Also, all these methods detect a fix set of local regions independently of the comparison process, and they typically fuse these local information in the last stage of similarity estimation. In contrast to aforementioned methods, our model introduces iterative recurrence to fuse the learned co-dependent features naturally into their similarity without spatial constraints.

More recently, one-shot learning is developed with deep neural networks for rapid knowledge acquisition \cite{Match-nets,Siamese-one-shot} and this idea has been applied into person re-ID \cite{GOG,One-shot-RE-ID,Cross-GANs} to address the problem of learning from few examples. Compared with these approaches, the proposed comparator is able to produce relative representations with respect to unseen examples on-the-fly, making our model more generalized.

\subsection{Deep Attention Models}

Some co-attention models are recently developed for word embedding \cite{Coattention,Bi-direction-attention} to capture the interactions between the question and the document. For example, Xiong et al \cite{Coattention} create the dynamic co-attention model which exploits the common deep learning technique of attention while encoding the context and question. This architecture computes an affinity matrix between each pair of words in the context and question, which is used to weight the continuous representations of the two documents. Our co-attention encoder is inspired by the dynamic co-attention model in terms of the affinity matrix between each feature grid to capture the interactions between paired images. However, the original co-attention model \cite{Coattention} doesn't have a fusion strategy for the detected co-dependent features, and thus not directly applicable to the task of person re-ID. The attentive recurrent comparators (ARC) \cite{ARC} is another effective model that employs a recurrent neural network with attention mechanism to compare two images by repeatedly cycling through both the images, attending one image at one time-step. However, in the model of ARC, the local finer common patterns across two images are not considered and the recurrent screening on two images is unable to memorize which sets of local patterns are critical to distinguish the two objects. In contrast, our model effectively detect and localize common patches via the co-dependency operation which provide insights on re-identifying persons.
\section{Deep Co-attention Comparators (DCCs) for Person Re-identification On-the-fly}\label{sec:approach}

In this section, we present the proposed deep co-attention comparator (DCC) for paired person images. The goal is to learn co-dependent features for paired input and their similarity metric simultaneously. The primary challenge is to effectively learn dynamic features from discriminative patches relative to the paired images, namely co-attention features, and jointly learn a similarity conditioned on both images.

\subsection{Convolutional Representations}

\begin{figure*}[t]
\centering
\begin{tabular}{c}
\includegraphics[height=2cm,width=14cm]{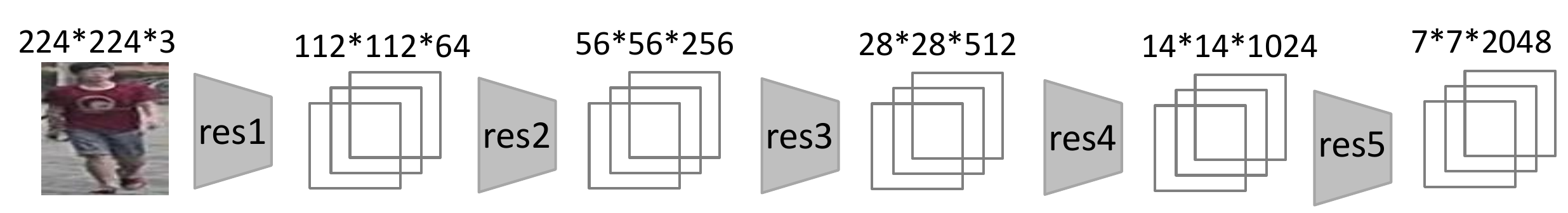}
\end{tabular}
\caption{The convolution with parameters of ResNet-50 model. We use the high-level feature maps from Res4 block as the base features to perform co-attention detection.}
\label{fig:ResNet}
\end{figure*}

Recent studies have shown that CNN models can be fine-tuned on person re-ID datasets to learn more discriminative visual appearance \cite{PIE-reid,PAN,DGGropout}. To retain the spatial region based features for each image, we use the convolutional blocks extracted from the ResNet-50 model \cite{ResNet} as the base model to produce base features for the co-attention detection. The ResNet-50 model consists of five down-sampling blocks and one global average pooling. Each residual block encapsulates several convolutional layers with batch normalization, ReLU, and optionally max pooling. After each block, the feature maps are down-sampled to be half-sized of the maps in the previous block. The design of ResNet is shown in Fig.\ref{fig:ResNet}.  We deploy the model pre-trained on the ImageNet and remove the fully-connected layers. Thus, the base features for each image are extracted from the fourth residual block of ResNet-50 without performing the average pooling operation, that is, the extracted base features are in the size of $14\times 14 \times 1024$.

\subsection{Deep Co-attention Encoder}\label{ssec:coattention}

\begin{figure*}[t]
\centering
\begin{tabular}{c}
\includegraphics[height=3cm]{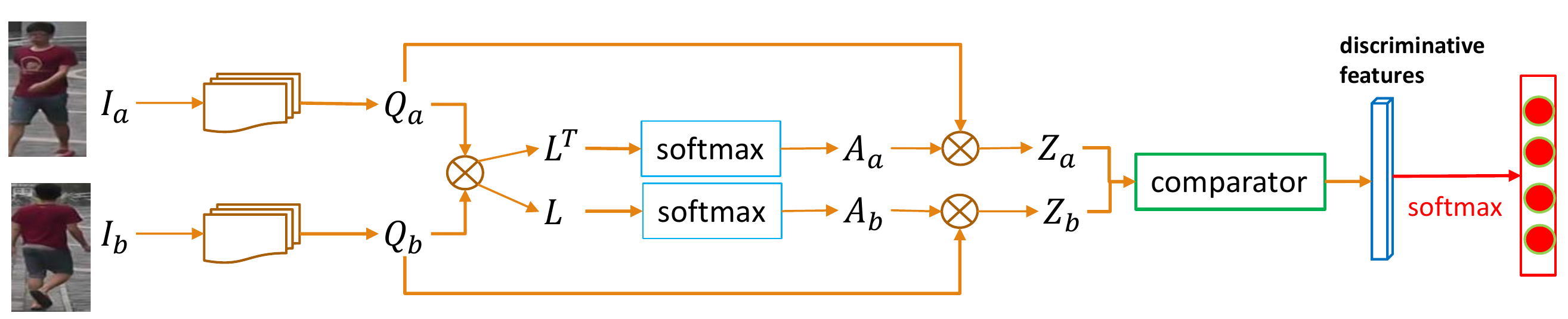}
\end{tabular}
\caption{The co-attention encoder over the paired input $I_a$ and $I_b$. The feature maps delivered by the ResNet $\bQ_a$ and $\bQ_b$ are used to compute the affinity matrix $\bL$ which can be normalized ($\bA_a$ and $\bA_b$) to compute the attention weight summaries w.r.t each other $\bZ_a$ and $\bZ_b$. Then, the comparator iteratively fuses $\bZ_a$ and $\bZ_b$ to produce discriminative features for the training objective. See the text for details.}
\label{fig:coattention}
\end{figure*}

The basic idea of co-attention encoder is illustrated in Fig.\ref{fig:coattention}. Given the feature activations $\bQ_a \in \mathbb{R}^{C\times M\times M}$ and $\bQ_b \in \mathbb{R}^{C\times M\times M}$ regarding the paired inputs $I_a$ and $I_b$, the co-attention encoder is to fuse the two-resource features co-dependently on each other. To do this, we first compute the affinity matrix
\begin{equation}\label{eq:affinity}
\bL= \bQ_b^T \bW^{(L)} \bQ_a \in \mathbb{R}^{M^2\times M^2},
\end{equation}
where $\bW^{(L)}\in \mathbb{R}^{M^2 \times M^2}$ is a trainable weight matrix to be learned through the networks. Specifically, each entry $(i,j)$ in $\bL$ represents the similarity between each feature $i$ in the gallery image $I_b$ and feature $j$ in the query image $I_a$. Then, the $\bL$ is normalized to produce attention weights for the respective representations of $I_a$ and $I_b$. We compute
\begin{equation}
\begin{split}
&\bA_a= softmax(\bL) \in \mathbb{R}^{M^2\times M^2};\\
&\bA_b=softmax(\bL^T) \in \mathbb{R}^{M^2\times M^2},
\end{split}
\end{equation}
where $softmax(\cdot)$ normalizes each row of the input. The $i$-th row of $\bA_a$ is a vector of length $M^2$ with weights describing the relevance of each feature in $\bQ_a$ to the $i$-th feature in $\bQ_b$. Similarly, the $j$-th row of $\bA_b$ is a vector of length $M^2$ with weights describing the relevance of each feature in $\bQ_b$ to the $j$-th feature in $\bQ_a$.

With the attention weights, the attention summaries for the query $\bQ_a$ with respect to the gallery $\bQ_b$ can be computed by
\begin{equation}\label{eq:attention_query}
\bZ_a= \bQ_b \bA_a \in \mathbb{R}^{C\times M^2}.
\end{equation}
In Eq.\eqref{eq:attention_query}, for each feature in the query image, we compute a weighted average over all the features in the gallery image where the weights are provided by the rows in $\bA_a$ corresponding to that feature in the gallery image. Likewise, the attention summaries for the gallery $\bQ_b$ with respect to the query can be computed by
\begin{equation}\label{eq:attention_gallery}
\bZ_b= \bQ_a \bA_b \in \mathbb{R}^{C\times M^2}.
\end{equation}

In the next section, we will detail the proposed iterative recurrence based comparator to fuse the co-dependent features to produce discriminant representations while relative to the input pair.

\subsection{Iterative Recurrence based Similarity Comparator: Fusing Co-dependent Features}\label{ssec:iterative-comparator}

\begin{figure}[t]
\centering
\begin{tabular}{c}
\includegraphics[height=4cm]{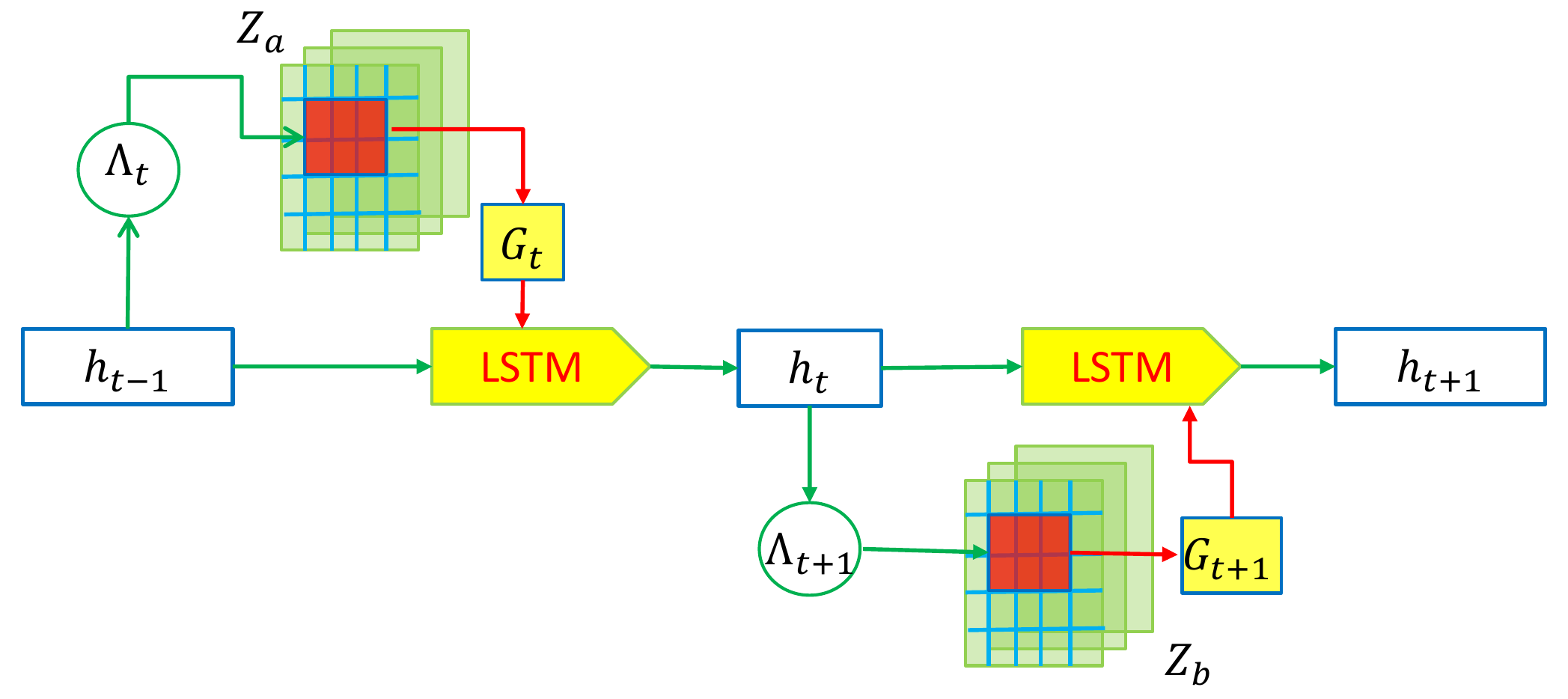}
\end{tabular}
\caption{The iterative fusion on co-dependent features. The comparator attends to distinct features $\bG_t$ (determined by attention glimpse $\Lambda_t$ and hidden state $\bh_{t-1}$) alternatively across two feature blocks $\bZ_a$ and $\bZ_b$ to fuse them into discriminative representations for the similarity learning.}
\label{fig:iterative-com}
\end{figure}

The comparator consists of a recurrent neural network and an attention mechanism that takes the co-dependent pair with a specially constructed presentation sequence. The comparing process is essentially an imitation of the human perception. The prominent characteristic of this process is to perceive new observations conditioned on the previous context that has been investigated so far by the observer. A series of such repeated observations are accumulated to form a final judgement on their similarity \cite{ARC}. Similar to the attentive recurrence comparator (ARC) \cite{ARC}, our co-dependent comparator is premised on recurrent neural networks as the backbone and an attention mechanism which alternates across two images while attending to the salient features. The overview of this comparator with fusion on co-dependent features are shown in Fig.\ref{fig:iterative-com}.

Given co-dependent features $\bZ_b$, $\bZ_b$ for respective images $I_a$ and $I_b$, the recurrent comparator carries out an iterative representation fusion from both images in a sequence of $\{\bZ_a \hookrightarrow \bZ_b \hookrightarrow \bZ_a \hookrightarrow \bZ_b, \ldots, \bZ_a \hookrightarrow \bZ_b \}$. For one time step $t$, the current image under observation is determined via
\begin{equation}\label{eq:image-at-now}
\bZ_t = \begin{cases} \bZ_a  &\mbox{if } t \% 2 = 0 \\
\bZ_b & \mbox{else } \end{cases}
\end{equation}

\begin{equation}\label{eq:glimpse}
  \bG_t = f_{att}(\Lambda_t , \bZ_t),
\end{equation}
where $f_{att}(\cdot)$ is the attention function that acts on the feature block $\bZ_t$ (The detail of $f_{att}$ will be described in Section \ref{sssec:attention}). $\Lambda_t=\bW_g \bh_{t-1}$ is the attention glimpse which specifies the location and size of an attention window. At each time step, the attention glimpse $\Lambda_t$ is computed by a projection from the previous hidden state of the RNN steam, that is, $\bh_{t-1}$ where the projection is parameterized by the $\bW_g$ that maps the $\bh_{t-1}$ to the trainable number of attention parameters. Then, the next hidden state $\bh_t$ can be sequentially computed by using the glimpse $\bG_t$ and the previous hidden state $\bh_{t-1}$ via
\begin{equation}
\bh_t= RNN(\bG_t, \bh_{t-1}),
\end{equation}
where the $RNN(\cdot)$ is the recurrence function which is implemented as an LSTM in this paper.

\subsubsection{The Attention Mechanism}\label{sssec:attention}

To endow the network with selective attention while retaining the benefits of gradient decent training, we take the two dimensional form of attention that is recently used in ``DRAW" \cite{DRAW}, where an array of 2D Gaussian filters is applied to the image, yielding an image patch of smoothly varying location and zoom. Let $K\times K$ be the grid of Gaussian filters which is positioned on the image by specifying the co-ordinates of the grid centre $(g_X, g_Y)$ and the stride distance $\delta$ between adjacent filters. The location of a $i$-th row, $j$-th column in the patch can be determined by the grid centre and stride as follows:

\begin{equation}\label{eq:grid}
\begin{split}
     & \mu_X^i= g_X + (i-K/2-0.5)/\delta; \\
     & \mu_Y^i= g_Y + (i-K/2-0.5)/\delta
\end{split}
\end{equation}

Hence, the grid's location and sized is defined based on the glimpse parameters. The $K\times K$ grid of kernels is placed at the central of $(g_X, g_Y)$ on the $A\times B$ image with the square in the grid has a length of $\delta$.

Given an image of size $A\times B$, the glimpse parameter set $\Lambda_t$ is unpacked to yield $(\hat{g_X},\hat{g_Y},\hat{\delta})$, and $g_X$, $g_Y$, $\delta$ can be computed from $(\hat{g_X},\hat{g_Y},\hat{\delta})$ via the following formulations:
\begin{equation}\label{eq:g-X-g-Y}
\begin{split}
     & g_X= (A-1) \frac{\hat{g_X} + 1}{2}; g_Y= (B-1) \frac{\hat{g_Y}+1}{2}; \\
     & \delta= \frac{\max(A,B)}{K-1} |\hat{\delta}|; \gamma=e^{1-2 |\hat{\delta}|}.
\end{split}
\end{equation}
where $\gamma$ is a scalar intensity that multiples the Gaussian filter response. The scaling of $g_X$, $g_Y$, and $\delta$ are chosen to ensure that the initial patch (with a randomly initialized network) roughly covers the whole input image.

\begin{figure}[t]
\centering
\begin{tabular}{c}
\includegraphics[height=4cm]{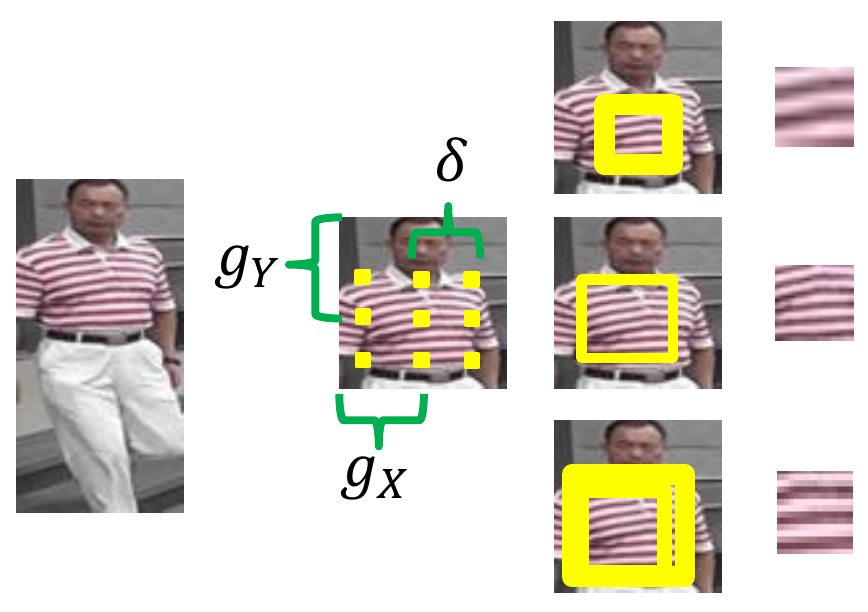}
\end{tabular}
\caption{The selective attention model. A $3\times 3$ filter super-imposed on the image with a stride $\delta$ and the centre location ($g_X$, $g_Y$). Three $K\times K$ patches extracted from the image ($K=12$). The top patch has a small $\delta$ and a high $\gamma$, the middle patch has large $\delta$ and low $\gamma$, and the bottom patch has large $\delta$ and high $\gamma$.}
\label{fig:selective-attention}
\end{figure}

The horizontal and vertical filter-bank matrices $F_X$ and $F_Y$ (dimensions are $K\times A$ and $K\times B$ respectively) are computed as
\begin{equation}\label{eq:filterbank}
\begin{split}
     & F_X [i,a]= \frac{1}{C_X} \{\pi \gamma [1+ (\frac{a-\mu_X^i}{\gamma})^2]\}^{-1};\\
     & F_Y[j,b]=\frac{1}{C_Y} \{\pi \gamma [1+ (\frac{a-\mu_Y^i}{\gamma})^2]\}^{-1}
\end{split}
\end{equation}
where $(i,j)$ is a point in the attention patch, $(a,b)$ is a point in the input image, and $C_X$ and $C_Y$ are normalization constants that ensure that $\sum_a F_X[i,a]=1$ and $\sum_b F_X[j,a]=1$. Therefore, the attention on an image can be computed via:
\begin{equation}\label{eq:attention}
  \bG_t=f_{att}(\Lambda_t,\bZ_t)=F_Y \bZ_t F_X^T.
\end{equation}
The illustration on the selective spatial attention mechanism is shown in Fig.\ref{fig:selective-attention}.

\subsection{The Similarity Learning and Loss Function}

Assume that we make $\mathcal{G}$ glimpses or observations jumping across paired images, and the hidden state of the RNN stream at the final time-step $\bh_{T}=\bh_{2\mathcal{G}}$ can be used as the relative representation of $I_a$ with respect to $I_b$ or vice versa.  To encode the similarity learning into the representation learning, we train the networks with an objective: a series of one-linear functions that map the input sample to a point in the representation space and a classifier that learns a decision boundary in this representation space. To this end, given a training batch containing samples from $\mathcal{C}$ classes (each class refers to a person identity), one sample is randomly selected as the unknown image and compare with another sample uniformly selected from $\mathcal{C}$ classes to form the paired input for training. Since the relative representations are the final hidden state of the LSTM for the training pair $I_u, I_j$ (where $I_u$ is the unknown image to be compared against each image $I_j$ with its known labels $j\in [1, \mathcal{C}]$), that is $\be_j =\bh_T (I_u, I_j)$, each embedding with respect to each class is mapped to a single score $s_j=f(\be_j)$. $f(\cdot)$ is an affine transformation function followed by a non-linearity. The final output is the normalized similarity with respect to all similarity scores:
\begin{equation}
p_j = softmax(W_j s_j), \forall j\in [1, \mathcal{C}].
\end{equation}
where $W_j \in \mathbb{R}^{1\times \mathcal{C}}$ is the weight vector. This softmax normalization allows for the relative similarity in the context of all training classes instead of an absolute similarity. The whole training procedure is differentiable and the gradient descent training can be applied.

\section{Experiments}\label{sec:exp}

\subsection{Data Sets}

\begin{figure}[t]
\centering
\begin{tabular}{c}
\includegraphics[height=3.5cm]{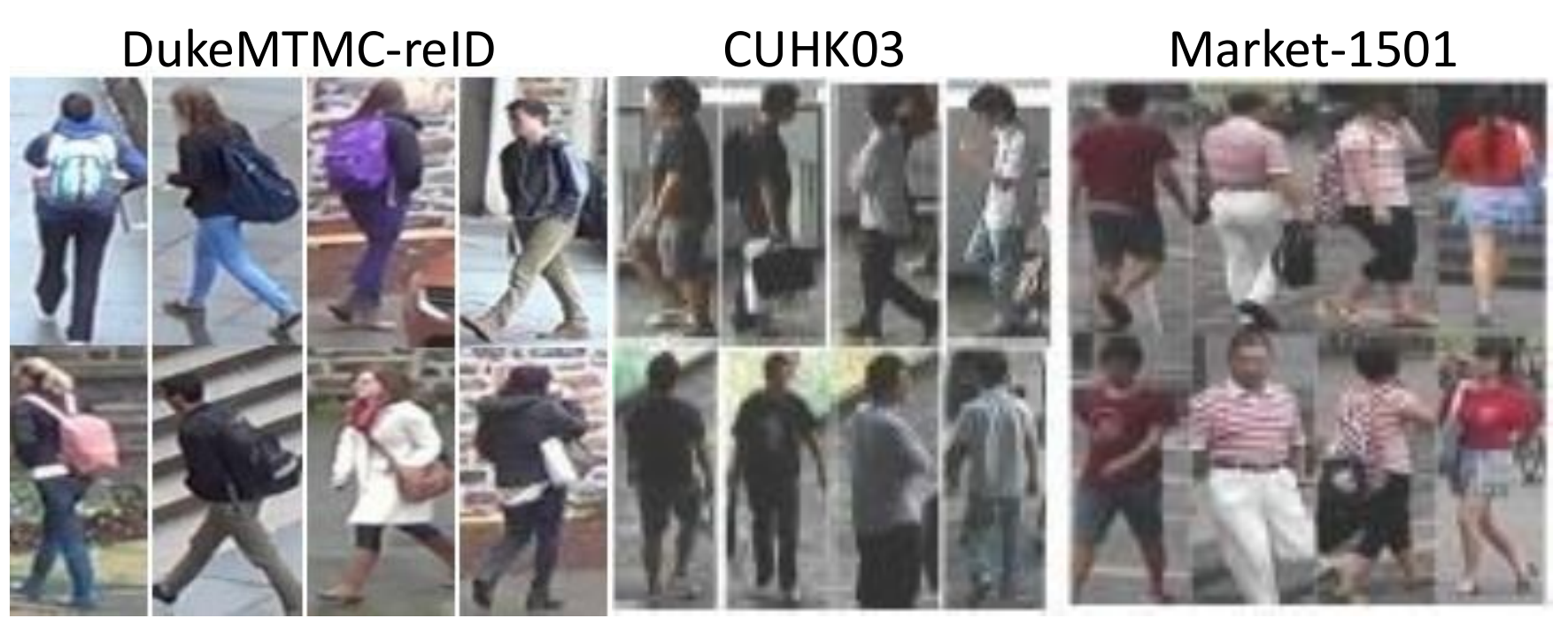}
\end{tabular}
\caption{Three person re-ID datasets.}
\label{fig:datasets}
\end{figure}

We perform experiments on three benchmarks for image-based person re-id: DukeMTMC-reID \cite{LSRO}, CUHK03 \cite{FPNN}, and Market-1501 \cite{Market1501}. Example images from the three benchmarks are shown in Fig.\ref{fig:datasets}.

\begin{itemize}
\item The \textbf{DukeMTMC-reID} dataset is a subset of the DukeMTMC \cite{DukeMTMC} for image-based person re-ID. The dataset is created from high-resolution videos from 8 different cameras. It is one of the largest pedestrian image datasets wherein images are cropped by hand-drawn bounding boxes. The dataset consists 16,522 training images of 702 identities, 2,228 query images of the other 702 identities and 17,661 gallery images. We follow the evaluation protocol in \cite{LSRO}.
\item The \textbf{CUHK03} dataset includes 13,164 images of 1360 pedestrians. It is captured with six surveillance cameras. Each identity is observed by two disjoint camera views, yielding an average 4.8 images in each view. We perform experiments on manually labeled dataset with pedestrian bounding boxes. The dataset is randomly partitioned into training, validation, and test with 1160, 100, and 100 identities, respectively.
 \item The \textbf{Market-1501} data set contains 32,643 fully annotated boxes of 1501 pedestrians, making it the largest person re-id dataset to date. Each identity is captured by at most six cameras and boxes of person are obtained by running a state-of-the-art detector, the Deformable Part Model (DPM) \cite{MarketDetector}. The dataset is randomly divided into training and testing sets, containing 750 and 751 identities, respectively.
\end{itemize}

\subsection{Experimental Settings}

We use the ResNet-50 model pre-trained on the ImageNet and extract the CNN feature maps from the fourth residual blocks. To train the model, we use mini-batch gradient descent with batch size of 128, and an initial learning rate of 0.001. In order to improve the training capability of our model in later epoches, we use an exponentially decaying learning rate with a decay rate of 0.88, which can be expressed as: $l=0.001 \times (0.88)^{m/N}$ where $m$ is the total number of mini-batches that have been processed for training and $N$ is the number of mini-batches in one epoch. The avoid the exploding gradient problem, we employ the gradient clipping when the sum of all gradient norms exceeds 100. For the optimizer, we use the ADAM algorithm \cite{ADAM} and train for 50 epoches. The units of hidden states are set to be 400 for LSTM. All the weights and biases are initialized using the Xavier initialization. To avoid over-fitting, we apply a dropout of 0.3 to the outputs of the LSTMs. The number of glimpse is set to be 8, thus making the total number of recurrent steps 16 for each paired input. Similar or dissimilar pairs of persons are randomly chosen to make the task more challenging. The attention window size is set to be $2\times 2$.

The evaluation protocol we adopt is the widely used single-shot modality to allow extensive comparison. Each probe image is matched against the gallery set, and the rank of the true match is obtained. The rank-$k$ recognition rate is the expectation of the matches at rank $k$, and the cumulative values of the recognition rate at all ranks are recorded as the one-trial Cumulative Matching Characteristic (CMC) results. This evaluation is performed ten times, and the average CMC results are reported.

\begin{figure}[t]
\centering
\begin{tabular}{c}
\includegraphics[height=4cm,width=8cm]{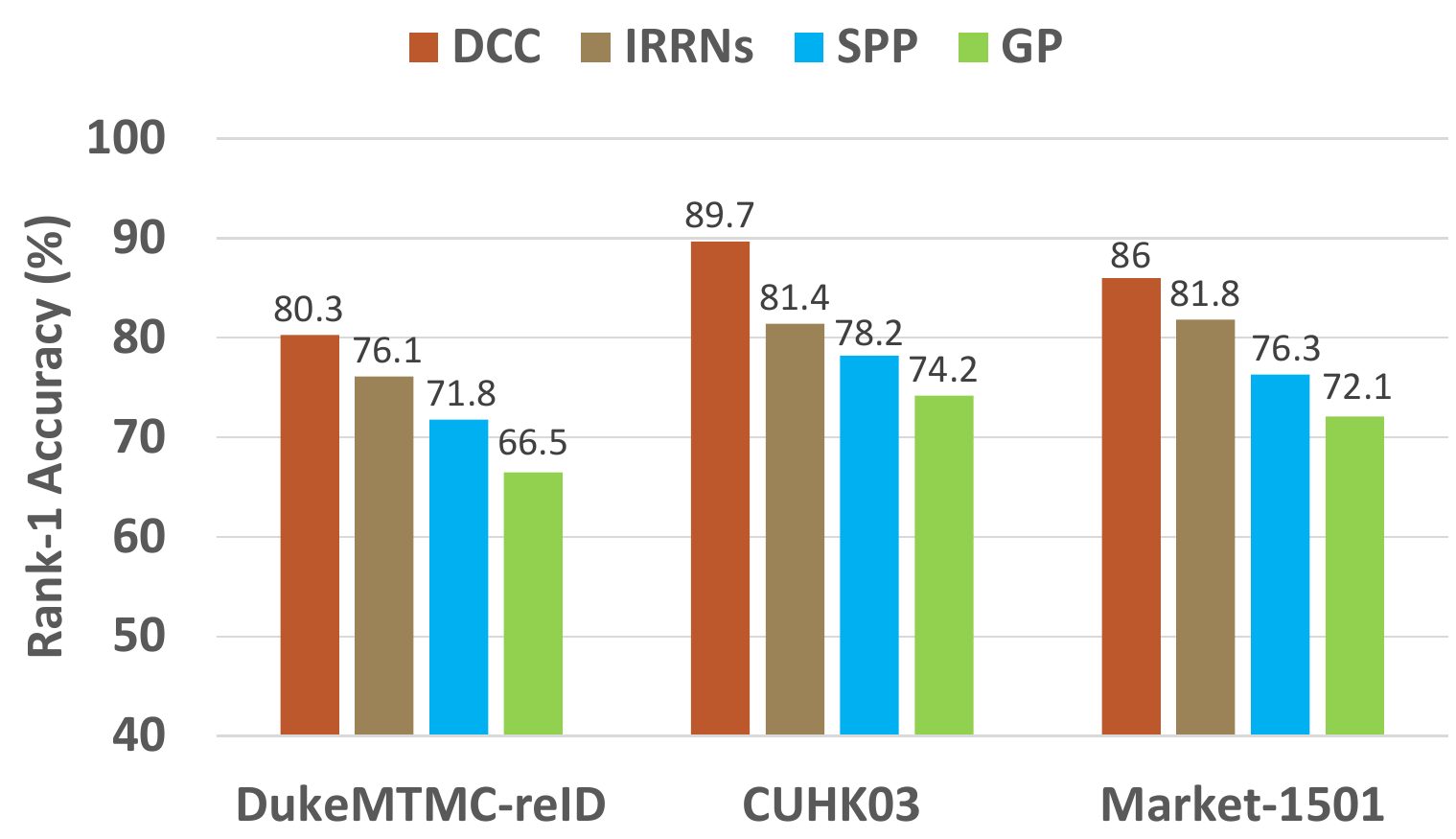}
\end{tabular}
\caption{The effect of iterative fusion based feature learning. See the texts for details.}
\label{fig:effect-iterative-fusion}
\end{figure}

\begin{figure*}[hbt]
\centering
\begin{tabular}{cc}
\includegraphics[height=4cm,width=9cm]{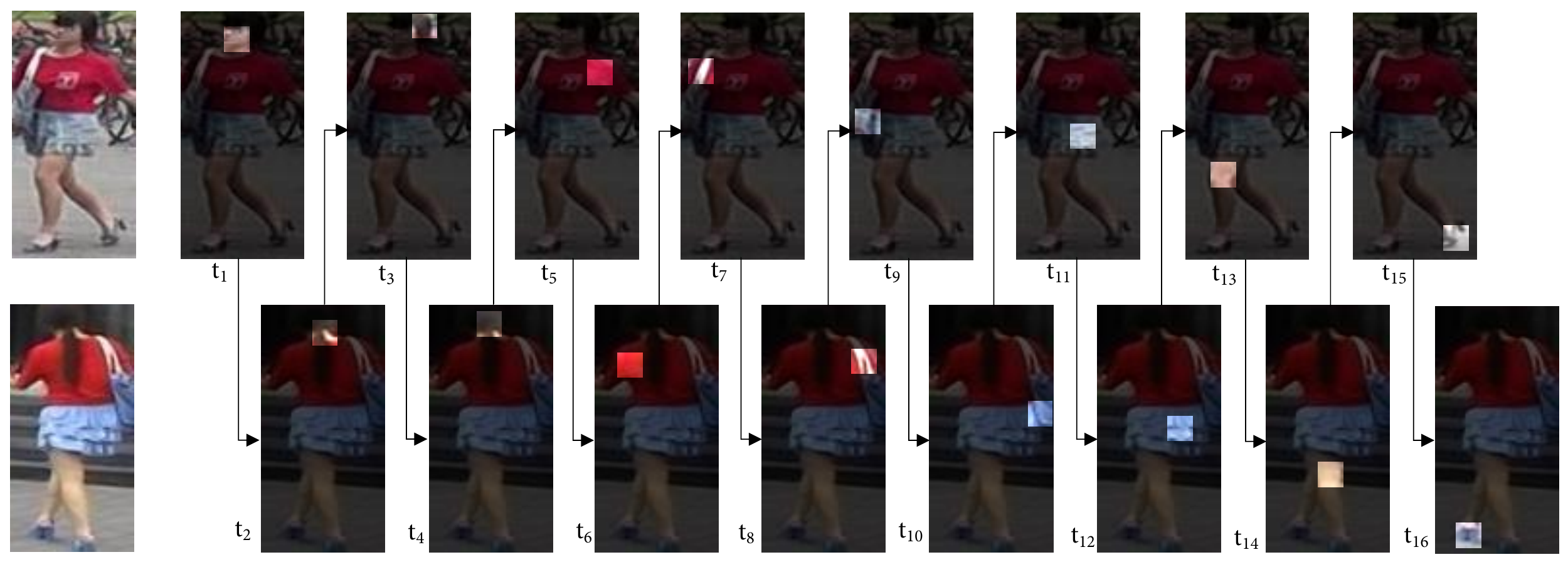}&
\includegraphics[height=4cm,width=9cm]{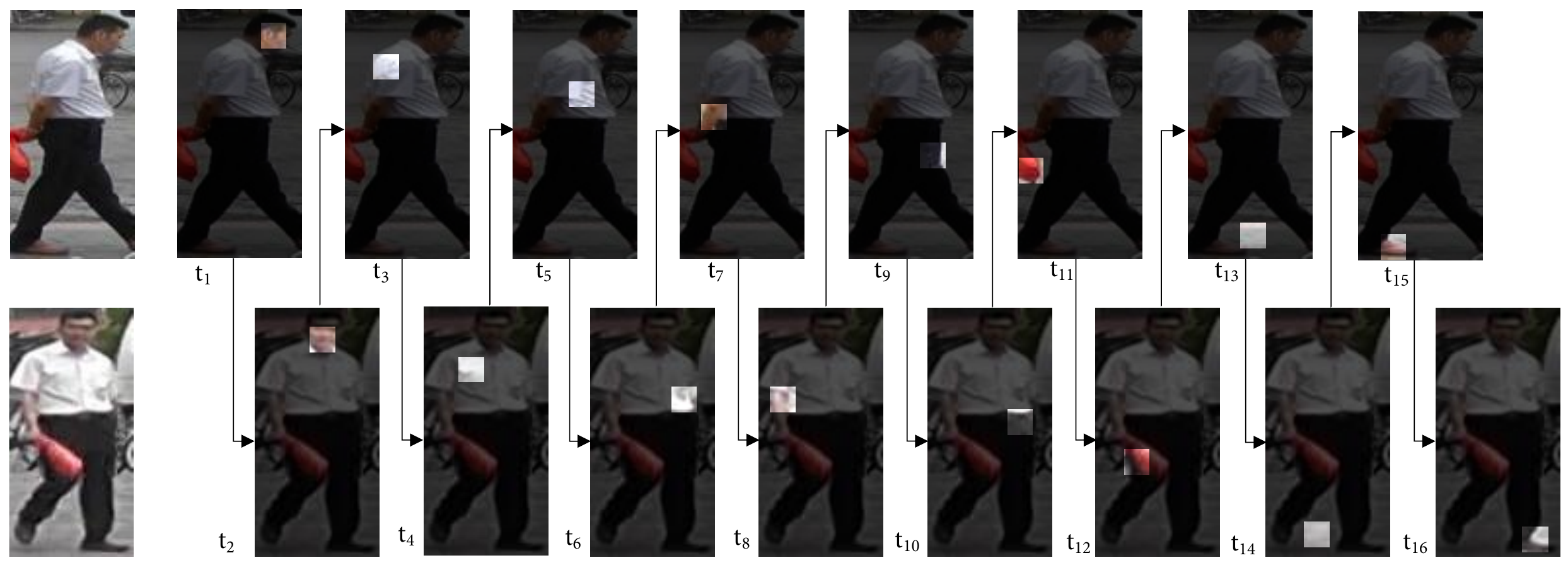}
\end{tabular}
\caption{The illustration on local regions attended by the recurrence comparator. For each pair of images, a succession of 16 glimpses ($t_1, \ldots, t_{16}$) taken by the DCCs alternates between both images to accumulate and fuse local regions into the final judgment of similarity.}
\label{fig:visualization-co-dependent}
\end{figure*}

\subsection{Ablation Studies}

In this section, we provide detailed analysis with insights to understand the ingredient contributions yielded by iterative fusion based feature learning and the generalization capability of the proposed method in similarity estimation.

\paragraph{The Effect of Iterative Fusion based Feature Learning}

The iterative fusion on co-dependent features is to resemble the human perception on object comparison which is incorporated into our model to avoid the spatial manipulation. This iterative fusion built on recurrence is a natural way to mimic human eyes in comparing objects and fusing informative patterns into the optimal similarity learning. Hence, to thoroughly study the contribution of the iterative fusion on co-dependent features, we conduct experiments by comparing to the alternative spatial encoding strategies.

\begin{itemize}
  \item Four-directional RNNs: Spatially recurrent pooling with IRNNs \cite{IRNN} have been used in \cite{What-and-where} to allow the spatial manipulation of feature maps from two CNN streams. In this spatial pooling pipeline, four RNNs sweep over the entire feature map in four directions: bottom to top, top to bottom, left to right, and right to left. In particular, the RNNs are composed of rectified linear units (ReLU) and the recurrent weight matrix is initialized to be identity matrix, namely IRNNs. Thus, on top of the co-dependent features, we replace the iterative fusion with four-directional IRNNs to produce the integrated features.
  \item Spatial Pyramid Pooling (SPP): A spatial pyramid pooling layer \cite{SPP-Net} can be added on top of the co-dependent features where spatial information is maintained by max-pooling in local spatial bins. Following \cite{What-and-where}, we perform a 2-level pyramid: $2\times 2$ and $1\times 1$ subdivisions over the resulting co-dependent features.
  \item Global Pooling (GP): The global pooling provides the information about the entire image and one could apply a global and unpool (tile or repeat spatially) back to the original feature map. In global average pooling, each cell in the output depends on the entire input with the same value repeated.
\end{itemize}

To have fair comparisons, the two CNN flows $\bZ_a$ and $\bZ_b$ are first fused using the bilinear pooling \cite{BilinearCNNs}, which are followed by the respective spatial pooling options, namely IRNNs, SPP, and GP. The comparison results are shown in Fig.\ref{fig:effect-iterative-fusion}. It can be seen that DCCs performs better than alternative spatial pooling operations consistently over three benchmarks, and thus can demonstrate the effectiveness of the employed iterative fusion. Although the bilinear pooled feature integration over $\bZ_a$ and $\bZ_b$ allows the feature-wise interaction by outer product of each vector, the resulting features are orderless and thus additionally spatial pooling operations alike IRNNs, SPP or GP are needed to make the comparison in a spatial context. The separation on fusion and spatial encoding is not well generalized to unseen examples which may require flexible representations relative to each testing pair.

\paragraph{Visualizations on Attending to Co-dependent Features}

\begin{figure}[t]
\centering
\begin{tabular}{c}
\includegraphics[height=4cm,width=7cm]{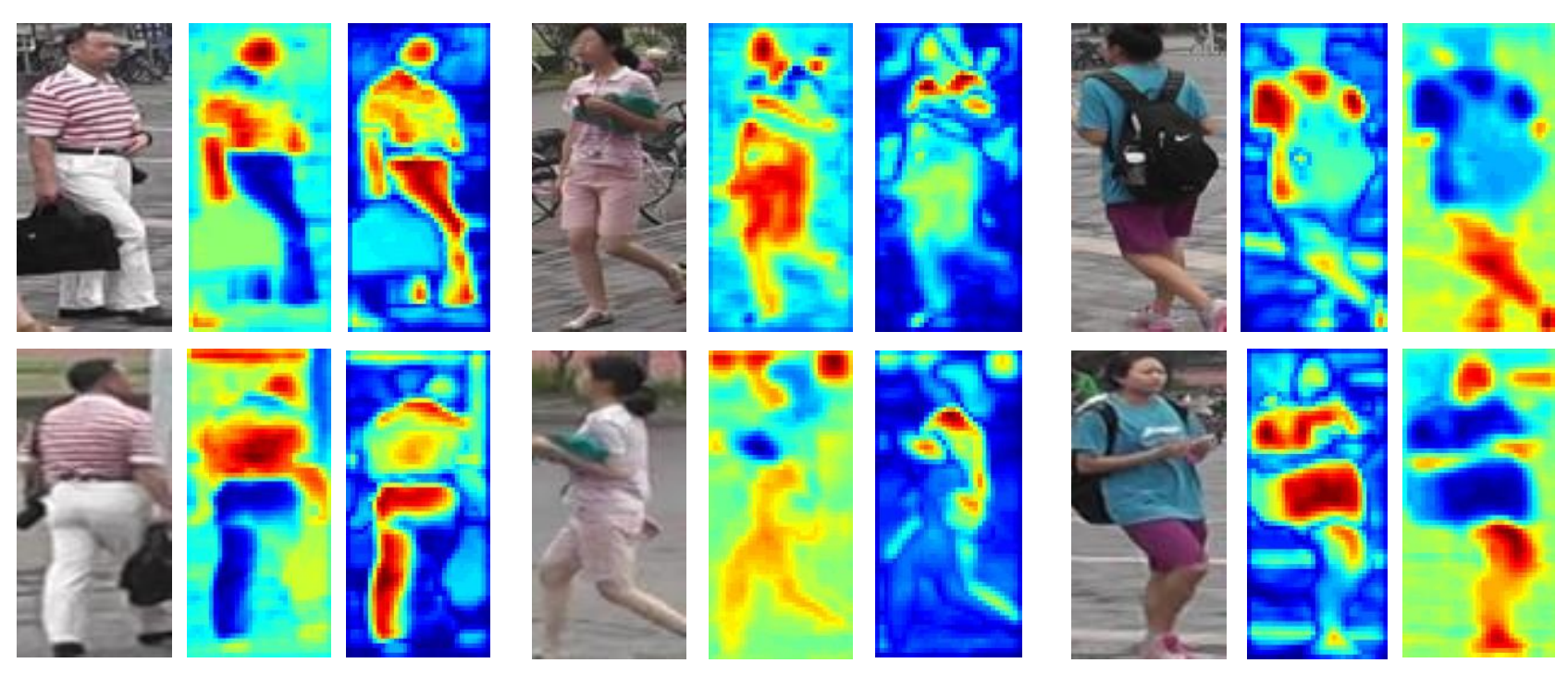}
\end{tabular}
\caption{The network is learned to effectively focus on concurrent regions across images.}
\label{fig:co-attention-maps}
\end{figure}

In this experiment, we illustrate the proposed iterative attention comparator that allows the network to alternatively jump across two images to attend on the relevant region of the image, and fuse extracted features into the final judgment of similarity. As shown in Fig.\ref{fig:visualization-co-dependent}, the recurrent comparator that receives a $12\times 12$ glimpse from a single image at each time-step by using the selective spatial attention defined in Section \ref{ssec:iterative-comparator}. After a sequence of glimpses, the most relevant regions are selected from the co-dependent features corresponding to the input images to be fused into the similarity learning. These features from detected regions are used in the subsequent softmax layer to compute the relative similarity w.r.t the training classes. For better illustration, in Fig.\ref{fig:co-attention-maps} we show the region features which are detected to be the concurrent on both images.

\paragraph{Comparison to One-shot/Few-shot Metric Learning Methods}

\begin{figure}[t]
\centering
\begin{tabular}{cc}
\includegraphics[height=4cm,width=4cm]{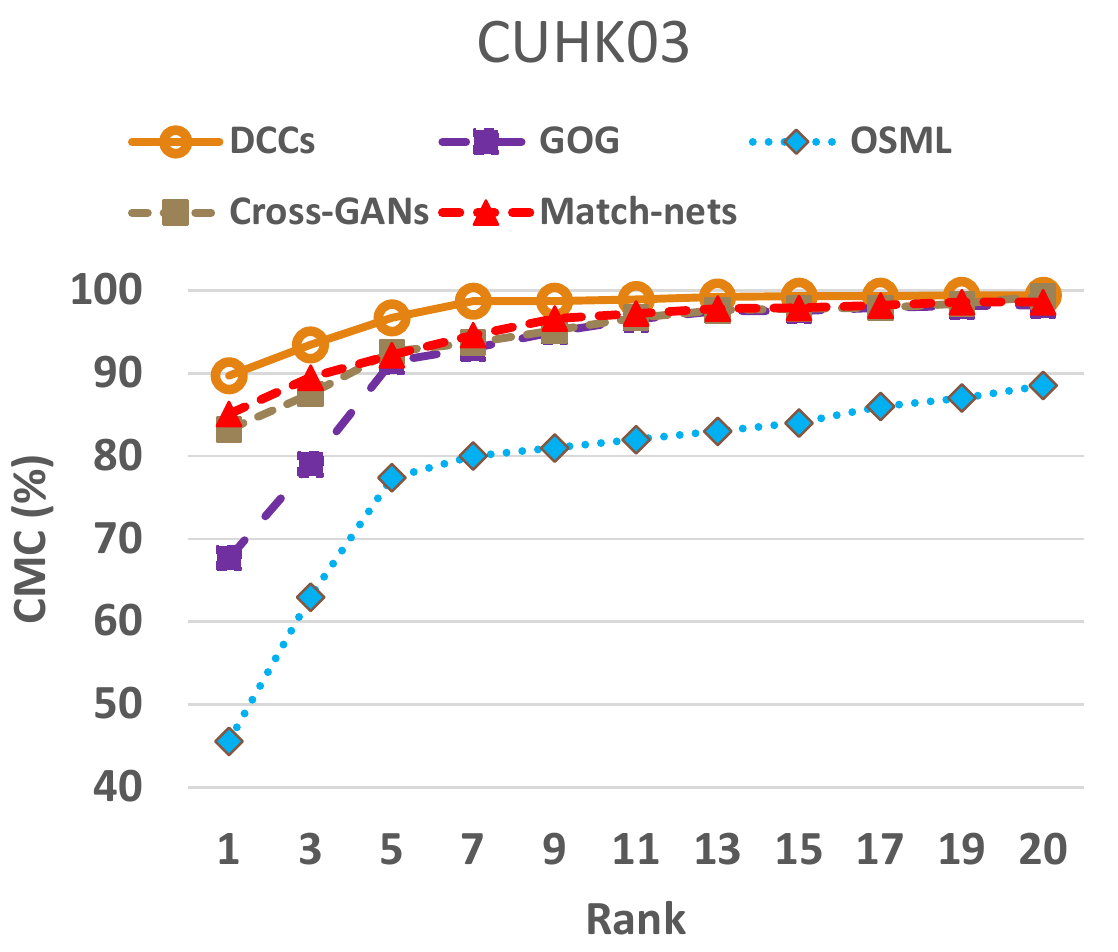}&
\includegraphics[height=4cm,width=4cm]{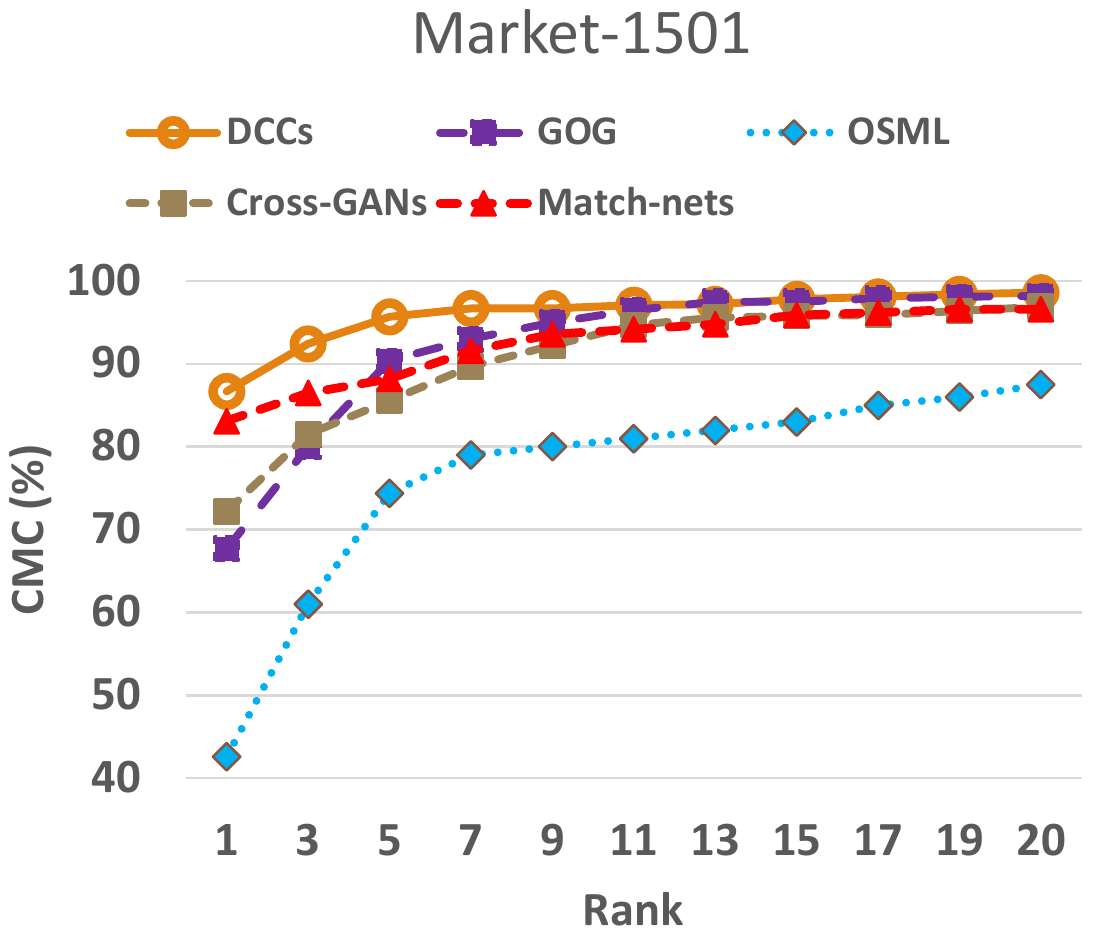}
\end{tabular}
\caption{The comparison to one/few-shot metric learning based sate-of-the-arts.}
\label{fig:compare-few-shot}
\end{figure}

\begin{table*}[hbt]
  \centering
  \caption{Rank-1, -5, -10, -20 recognition rate of one/few-shot metric learning methods on the CUHK03 and Market-1501 data sets.}\label{tab:few-shot-compare}
  {
  \begin{tabular}{l|c|c|c|c|c|c|c|c}
  \hline\hline
& \multicolumn{4}{c|}{CUHK03} & \multicolumn{4}{c}{Market-1501}\\
\cline{2-9}
    Method  & $R=1$  &  $R=5$ & $R=10$  & $R=20$ & $R=1$  &  $R=5$ & $R=10$  & $R=20$ \\
  \hline
   GOG \cite{GOG} & 67.7 & 91.4  & 96.5 & 98.1 & 65.8 & 90.4 & 96.5 & 98.2\\
   OSML \cite{One-shot-RE-ID} & 45.6 & 77.4 & 82.0 & 88.5 & 42.6 & 74.4 & 81.0 & 87.5\\
   Cross-GANs \cite{Cross-GANs} & 83.2 & 92.6 & 96.7 & 99.3 & 72.2 & 85.6 & 94.7 & 96.9 \\
   Match-nets \cite{Match-nets} & 85.1 & 92.2 & 97.2 & 98.6 & 83.1 & 88.2 & 94.2 & 96.6  \\
  \hline
   DCCs & \color{red}$\mathbf{89.7}$ & \color{red}$\mathbf{96.7}$ & \color{red}$\mathbf{98.9}$ & \color{red}$\mathbf{99.4}$ & \color{red}$\mathbf{86.7}$ & \color{red}$\mathbf{95.7}$ & \color{red}$\mathbf{97.1}$ & \color{red}$\mathbf{98.6}$ \\
  \hline
  \end{tabular}
  }
\end{table*}

In this experiment, we study the generalization of the proposed DCCs in similarity estimation by comparing with recent one/few-shot metric learning methods including GOG \cite{GOG}, OSML \cite{One-shot-RE-ID}, Cross-GANs \cite{Cross-GANs}, and Match-nets \cite{Match-nets}. The comparison results are reported in Table \ref{tab:few-shot-compare} and Fig.\ref{fig:compare-few-shot}. Compared with GOG \cite{GOG} and OSML \cite{One-shot-RE-ID} that carefully design hand-crafted features like color histogram and texture to be generalized to unseen examples, our DCCs outperform these methods by a large margin by 22.0\% and 21.2\% against GOG \cite{GOG} on CUHK03 and Market-1501, respectively. The main reason is that they perform feature engineering and metric learning separately, and thus unable to produce optimal features and/or metrics. Compared with the more compelling method of Cross-GANs \cite{Cross-GANs} that is to learn a joint distribution for cross-image representations by studying the co-occurrence statistic patterns, our method is superior to Cross-GANs \cite{Cross-GANs} by early fusing the feature learning into similarity learning from the beginning.

\subsection{Comparison to State-of-the-arts}

\begin{figure}[t]
\centering
\begin{tabular}{c}
\includegraphics[height=5cm,width=7cm]{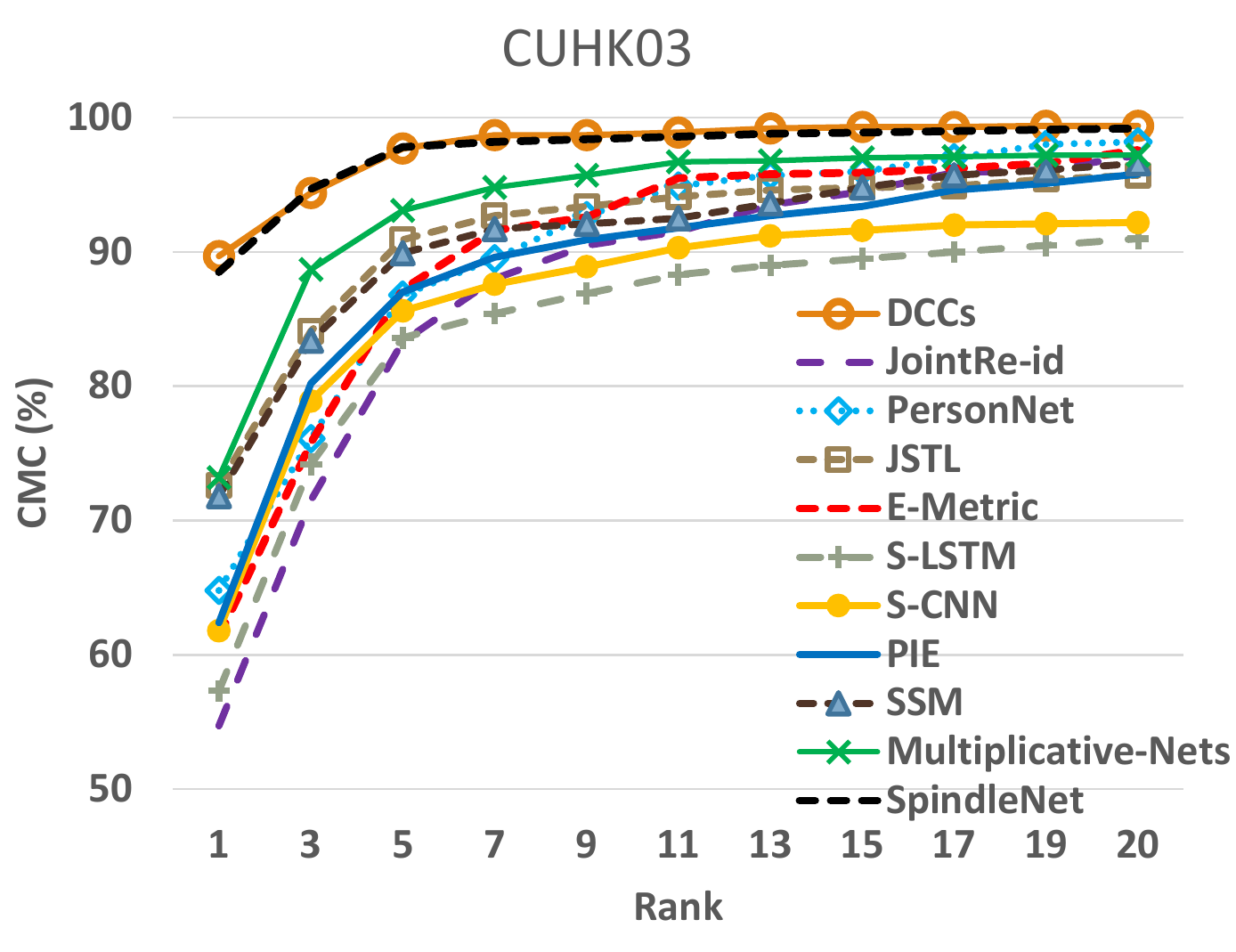}\\
\includegraphics[height=5cm,width=7cm]{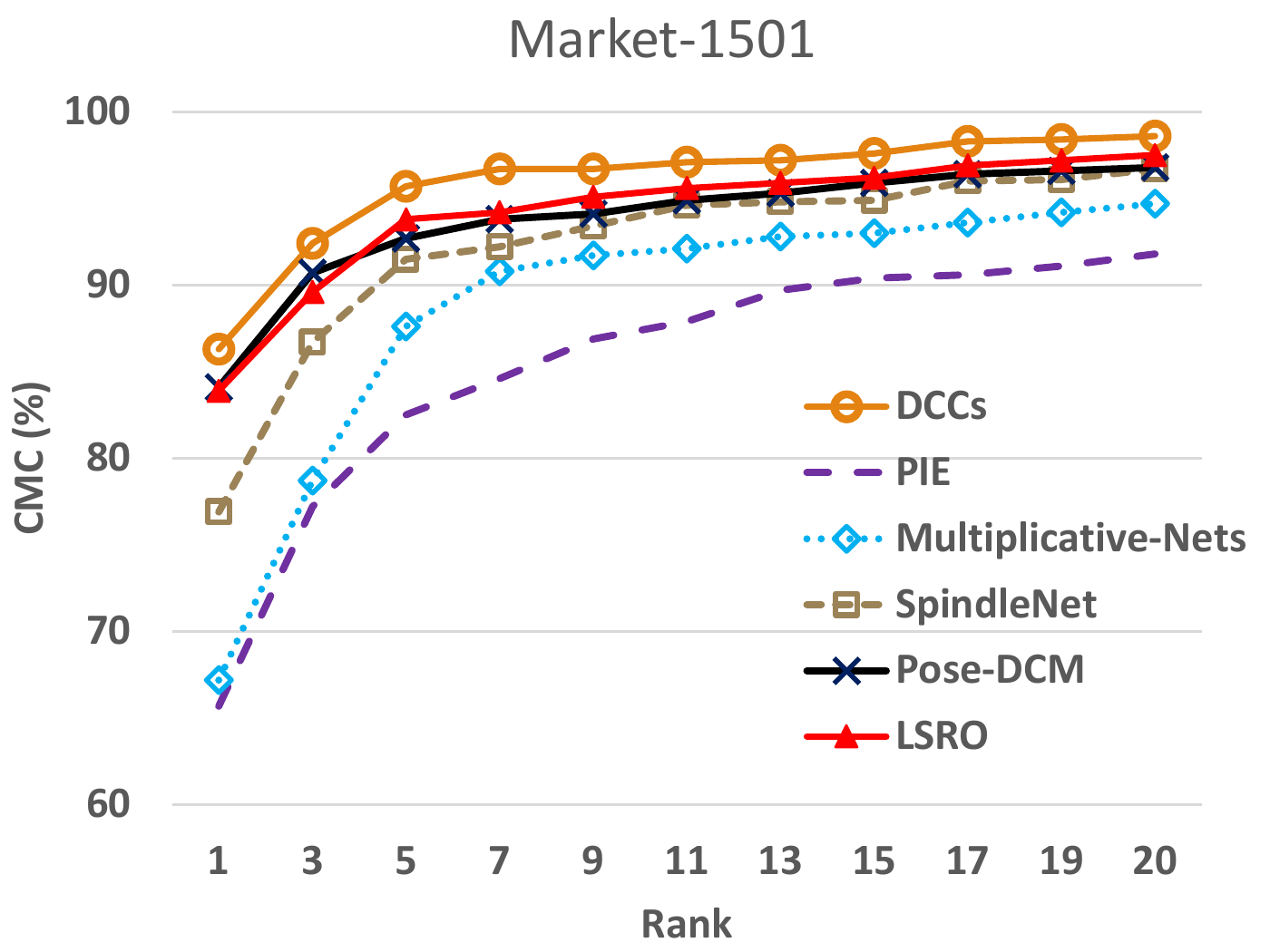}
\end{tabular}
\caption{The CMC curves on CUHK03 and Market-1501 datasets.}
\label{fig:cmc-cuhk03-market}
\end{figure}

\begin{table}[t]
  \centering
  \caption{Rank-1, -5, -10, -20 recognition rate and mAP of various methods on the DukeMTMC-reID data set.}  \label{tab:cmc-duke}
  {
  \begin{tabular}{l|c|c|c|c|c|c}
  \hline
  \multicolumn{2}{|c|}{\cellcolor{gray} \textcolor{white} {Method}} & \cellcolor{gray} \textcolor{white} {R=1} & \cellcolor{gray} \textcolor{white} {R=5} & \cellcolor{gray} \textcolor{white} {R=10} & \cellcolor{gray} \textcolor{white} {R=20} & mAP\\
  \hline
   \parbox[t]{1mm}{\multirow{7}{*}{\rotatebox[origin=c]{90}{Semi/un-supervised}}} & DCCs & \color{red}$\mathbf{80.3}$ &  \color{red}$\mathbf{92.0}$ & \color{red}$\mathbf{97.1}$ & \color{red}$\mathbf{98.6}$ & \color{red}$\mathbf{59.2}$ \\
   \cline{2-7}
   & GOG \cite{GOG} & 65.8 & 90.4 & 96.5 & 98.2 & -\\
   & SPGAN \cite{SPGAN} & 46.4 & 62.3 & 68.0 & 73.8 & 26.2\\
   & XQDA \cite{LOMOMetric} & 30.8 & - &-&-& 17.0 \\
   & UMDL \cite{UMDL} & 18.5 & 31.4 & 37.6 &- & 7.3\\
   & PUL \cite{PUL} & 30.0 & 43.4 & 48.5 &- & 16.4\\
   & BoW \cite{Market1501} & 17.1 & 28.8 & 34.9 &- & 8.3\\
  \hline
  \parbox[t]{1mm}{\multirow{4}{*}{\rotatebox[origin=c]{90}{Supervised}}} & SVDNet \cite{SVDNet} & 76.7 &- &- &- & 56.8 \\
   & IDE \cite{IDE} & 66.7 & 79.1 & 83.8 & 88.7 & 46.3\\
   & PAN \cite{PAN} & 71.6 &- & -&- & 51.5 \\
   & LSRO \cite{LSRO} & 67.7 & -&- &- & 47.1\\
  \hline
  \end{tabular}
  }
\end{table}

\begin{table}[t]
  \centering
  \caption{Rank-1, -5, -10, -20 recognition rate of various methods on the CUHK03 data set.}  \label{tab:cmc-cuhk03}
  {
  \begin{tabular}{l|c|c|c|c|c}
  \hline
  \multicolumn{2}{|c|}{\cellcolor{gray} \textcolor{white} {Method}} & \cellcolor{gray} \textcolor{white} {R=1} & \cellcolor{gray} \textcolor{white} {R=5} & \cellcolor{gray} \textcolor{white} {R=10} & \cellcolor{gray} \textcolor{white} {R=20} \\
  \hline
     \parbox[t]{1mm}{\multirow{7}{*}{\rotatebox[origin=c]{90}{Semi/un-supervised}}} & DCCs & \color{red}$\mathbf{89.7}$ & 96.7 & \color{red}$\mathbf{98.9}$  & \color{red}$\mathbf{99.4}$  \\
   \cline{2-6}
   & eSDC \cite{eSDC} & 8.7 & 17.6 &38.3 & 53.4\\
   & XQDA \cite{LOMOMetric} & 52.2 & 80.4 & 92.1 & 96.2\\
   & UMDL \cite{UMDL} & 1.6 & 5.4 & 7.9 & 10.2\\
   & CAMEL \cite{CAMEL} & 31.9 & 54.6 & 68.1 & 80.6 \\
   & OSML \cite{One-shot-RE-ID} & 45.6 & 77.4 & 82.0 & 88.5 \\
   & GOG \cite{GOG} & 67.7 & 91.4  & 96.5 & 98.1 \\
  \hline
  \parbox[t]{1mm}{\multirow{10}{*}{\rotatebox[origin=c]{90}{Supervised}}} & JointRe-id \cite{JointRe-id} & 54.7 & 85.3 & 91.5 & 97.3 \\
  & PersonNet \cite{PersonNet} & 64.8 & 86.8 & 94.9 & 98.2 \\
  & JSTL \cite{DGGropout} & 72.6 & 90.9 & 93.5 & 96.7 \\
  & E-Metric \cite{E-Metric} & 61.3 & 87.2 & 95.5 & 97.5 \\
  & S-LSTM \cite{S-LSTM} & 57.3 & 83.6 & 88.3 & 91.0\\
  & S-CNN \cite{GatedCNN} & 61.8 & 85.6 & 89.3 & 92.2\\
  & PIE \cite{PIE-reid} & 62.4 & 87.0 & 91.8 & 95.8\\
  & SSM \cite{SSM} & 71.8 & 89.9 & 92.5 & 96.6\\
  & SpindleNet \cite{SpindleNet} & 88.5 & 97.8 & 98.6 & 99.2\\
  & Multiplicative-Nets \cite{What-and-where} & 73.2 & 93.1 & 96.7 & 97.2\\
  \hline
  \end{tabular}
  }
\end{table}

\begin{table}[t]
  \centering
  \caption{Rank-1, -5, -10, -20 recognition rate and mAP of various methods on the Market-1501 data set.}  \label{tab:cmc-market}
  {
  \begin{tabular}{l|c|c|c|c|c|c}
  \hline
  \multicolumn{2}{|c|}{\cellcolor{gray} \textcolor{white} {Method}} & \cellcolor{gray} \textcolor{white} {R=1} & \cellcolor{gray} \textcolor{white} {R=5} & \cellcolor{gray} \textcolor{white} {R=10} & \cellcolor{gray} \textcolor{white} {R=20} & mAP\\
  \hline
   \parbox[t]{1mm}{\multirow{9}{*}{\rotatebox[origin=c]{90}{Semi/un-supervised}}} & DCCs & \color{red}$\mathbf{86.7}$ & \color{red}$\mathbf{95.7}$ & \color{red}$\mathbf{97.1}$ & \color{red}$\mathbf{98.6}$ & \color{red}$\mathbf{69.4}$ \\
   & DCCs + KISSME \cite{KISSME} & \color{red}$\mathbf{88.4}$ & \color{red}$\mathbf{96.1}$ & \color{red}$\mathbf{97.5}$ & \color{red}$\mathbf{99.0}$ & \color{red}$\mathbf{71.1}$ \\
   \cline{2-7}
   & eSDC \cite{eSDC} & 33.5 & -&- &- & 13.5\\
   & SPGAN \cite{SPGAN} &  57.5 & 75.8 & 82.4 & 87.6 & 26.7\\
   & XQDA \cite{LOMOMetric} & 43.8 & - &-&-& 22.2 \\
   & UMDL \cite{UMDL} & 34.5 & 52.6 & 61.7 & 68.0 &-\\
   & PUL \cite{PUL} & 45.5 & 60.7 & 64.2 & 72.6  & -\\
   & BoW \cite{Market1501} &  34.4 & 61.5 & 72.8 & 82.5 & 14.1 \\
   & OSML \cite{One-shot-RE-ID} & 42.6 & 74.4 & 81.0 & 87.5 & -\\
  \hline
  \parbox[t]{1mm}{\multirow{12}{*}{\rotatebox[origin=c]{90}{Supervised}}} & S-CNN \cite{GatedCNN} & 65.9 &- &- &- & 39.6\\
   & PIE \cite{PIE-reid} & 65.7 & 82.5 & 87.9 & 91.6& 41.1 \\
   & SSM \cite{SSM} & 82.2 & - &- &- & 68.8 \\
   & SpindleNet \cite{SpindleNet} & 76.9 & 91.5 & 94.6 & 96.7 & -\\
   & Part-aligned \cite{Part-Aligned} & 81.0 & -& -& -&- \\
   & SVDNet \cite{SVDNet} & 82.3 &- &- & -&62.1 \\
   & Pose-DCM \cite{Pose-DCM} & 84.1 & 92.7 & 94.9 & 96.8 & 63.4\\
   & LSRO \cite{LSRO} & 83.9 & 93.8 & 95.6 & 97.5 & 66.1 \\
   & MSCAN \cite{MSCAN} & 80.3 &- & -&- & 57.5\\
   & Multiplicative-Nets \cite{What-and-where} & 67.2 & 87.6 & 92.1 & 94.7 & 40.2\\
\hline
  \end{tabular}
  }
\end{table}

In this section, we compare DCCs with state-of-the-arts including several unsupervised and semi-supervised approaches: eSDC \cite{eSDC}, CAMEL \cite{CAMEL}, SPGAN \cite{SPGAN}, UMDL \cite{UMDL}, OL-MANS \cite{OL-MANS}, PUL \cite{PUL}. Also, we consider some supervised competitors: S-CNN \cite{GatedCNN}, S-LSTM \cite{S-LSTM}, SSM \cite{SSM}, PIE \cite{PIE-reid}, SpindleNet \cite{SpindleNet}, Part-aligned \cite{Part-Aligned}, SVDNet \cite{SVDNet}, Pose-DCM \cite{Pose-DCM}, LSRO \cite{LSRO}, MSCAN \cite{MSCAN}. Note that not all these methods have their recognition values on the three data sets.

\paragraph{Experimental results on DukeMTMC-reID}

The experimental results on the DukeMTMC-reID dataset are reported in Table \ref{tab:cmc-duke}. It can be seen that comparing with semi/unsupervised methods, DCCs outperforms these baselines by a large margin. For instance, DCCs outperforms GOG \cite{GOG} by 14.5\% at rank-1 rate. The main reason is that most of semi/unsupervised methods perform feature extraction and metric learning in a separate manner while DCCs is to fuse feature learning effectively into similarity learning. Compared with other unsupervised methods of UMDL \cite{UMDL} and PUL \cite{PUL} which are based on transfer learning to transfer view-invariant representations learned from source datasets, our method achieves a significant performance gain by learning data-dependent features which can also be generalized into unseen observations. Compared with the most competing method of SVDNet \cite{SVDNet} that seeks feature representations with orthogonality, the proposed method leads to the performance improvement by 3.3\% at rank-1 rate and 2.4\% at mAP value.

\paragraph{Experimental results on CUHK03}

Table \ref{tab:cmc-cuhk03} reports the comparison results of our method and state-of-the-arts. The comparison results against semi/un-supervised methods are very similar as can be observed in the DukeMTMC-reID dataset. In comparison with supervised competitors, such as SpindleNet \cite{SpindleNet} that relies on body regions alignment, the proposed DCCs achieves better recognition result due to its more natural way of learning relative representations w.r.t paired images and its improved generalization. In comparison with our recent work of Multiplicative-Nets \cite{What-and-where} that rely on the spatial positioning of local features, the proposed DCCs can produce more flexible representations and relative to the whole training classes.

\paragraph{Experimental results on Market-1501}

The Market-1501 is the largest yet challenging dataset for image based person re-ID. The dramatic visual variations across unknown camera pairs require a highly flexible model in producing discriminative features while relative to the whole training set to make the features more generalized. The comparison results against state-of-the-arts in single query setting are shown in Table \ref{tab:cmc-market}. It can be seen that DCCs outperform all supervised methods and semi/un-supervised methods by achieving the state-of-the-art recognition rate of rank-1=86.7 and mAP=69.4. Additionally, the learned features in combination with a metric learning algorithm, that is, DCCs+KISSME \cite{KISSME} can further improve the rank-1=88.4, which is the current best accuracy in single query. In particular, compared with part-based approaches such as SpindleNet \cite{SpindleNet}, Part-aligned \cite{Part-Aligned}, and MSCAN \cite{MSCAN}, DCCs aim to learn features co-dependent on both images in comparison instead of single images, and thus for suitable for person verification.

\section{Conclusions and Future Work}\label{sec:con}

In this paper, we present the Deep Co-attention based Comparators to fuse the concurrent parts of both input images and produce their relative presentations. The proposed model is a more generalized deep neural network for person re-identification on-the-fly that is trained to be capable of rapidly focusing on relevant parts and estimating their similarity simultaneously. The DCCs consist of a co-attention encoder which learns co-dependent features w.r.t input pairs, and a recurrence comparator which iteratively cycles through paired images and estimate their similarity. We show that the proposed DCCs achieve much better performance than existing deep similarity learning methods in person re-ID. A possible extension of this model is to address the sequential scanning of the recurrence comparator which is computationally expensive.


\ifCLASSOPTIONcaptionsoff
  \newpage
\fi



\bibliographystyle{IEEEtran}\small
\bibliography{allbib}

\end{document}